\documentclass[10pt, conference, letterpaper]{IEEEtran}
\usepackage{cite}
\usepackage[caption=false,font=footnotesize]{subfig}
\usepackage{url}
\usepackage[utf8]{inputenc} 
\usepackage[USenglish,UKenglish,english]{babel} 
\usepackage[T1]{fontenc}    
\makeatletter
\newcommand{\upquotetype}{}
\newcommand{\upquote@aux}[1]{\text{\upquotetype}#1\text{\upquotetype}}
\newcommand{\upquotesingle}{\renewcommand{\upquotetype}{\textquotesingle}\upquote@aux}
\newcommand{\upquotedouble}{\renewcommand{\upquotetype}{\textquotedbl}\upquote@aux}

\usepackage{amsthm}

\usepackage{textcomp}

\usepackage{graphicx} 
\usepackage{epstopdf}
\usepackage{amsfonts}
\usepackage{amsmath}

\usepackage{algorithm}
\usepackage{algorithmic}

\usepackage{booktabs}
\newcommand{\mytab}{
  \begin{tabular}[b]{cc}
   & $\Delta RLE$ \\
\hline \\
MLP    & $-$ \\
TDRN   & $15.27\%$ \\
LRCN   & $18.53\%$ \\
SCCN   & $19.76\%$ \\
\\
  \end{tabular}
}

\begin{document}
\title{Summarized Network Behavior Prediction}

\author{\IEEEauthorblockN{Shih-Chieh Su\IEEEauthorrefmark{1}
}
\IEEEauthorblockA{
Qualcomm Inc.\\
San Diego, CA, 92121\\
Email: \IEEEauthorrefmark{1}shihchie@qualcomm.com}}

\maketitle

\begin{abstract}
This work studies the entity-wise topical behavior from massive network logs. Both the temporal and the spatial relationships of the behavior are explored with the learning architectures combing the recurrent neural network (RNN) and the convolutional neural network (CNN). To make the behavioral data appropriate for the spatial learning in CNN, several reduction steps are taken to form the topical metrics and place them homogeneously like pixels in the images. The experimental result shows both the temporal- and the spatial- gains when compared to a multilayer perceptron (MLP) network. A new learning framework called spatially connected convolutional networks (SCCN) is introduced to more efficiently predict the behavior.
\end{abstract}

\IEEEpeerreviewmaketitle

\section{Introduction}\label{sec-intro}

Understanding and predicting the behavior of an entity over a large domain of different actions is a challenging problem. The problem is even more difficult when the behavioral data is massively collected with lots of noise. There are various studies in using behavioral data as a global indicator. For instance, large scale activity data is used to measure and track the user experience, to decide the advertisement to be delivered, or to study the factors affecting the behavior, which can be applied to improve advertisement targeting. However, the aforementioned large scale behavioral analytics use cases have one aspect in common: they heavily simplified the response domain to have one or few learnable targets. 

\begin{figure}[!h]
\vskip -0.05in
  \centering
  \subfloat[Topical behavior of an entity. The same grid position over time is the metric of the same topic. Color means the intensity of the topical metric.]{\includegraphics[scale=0.055]{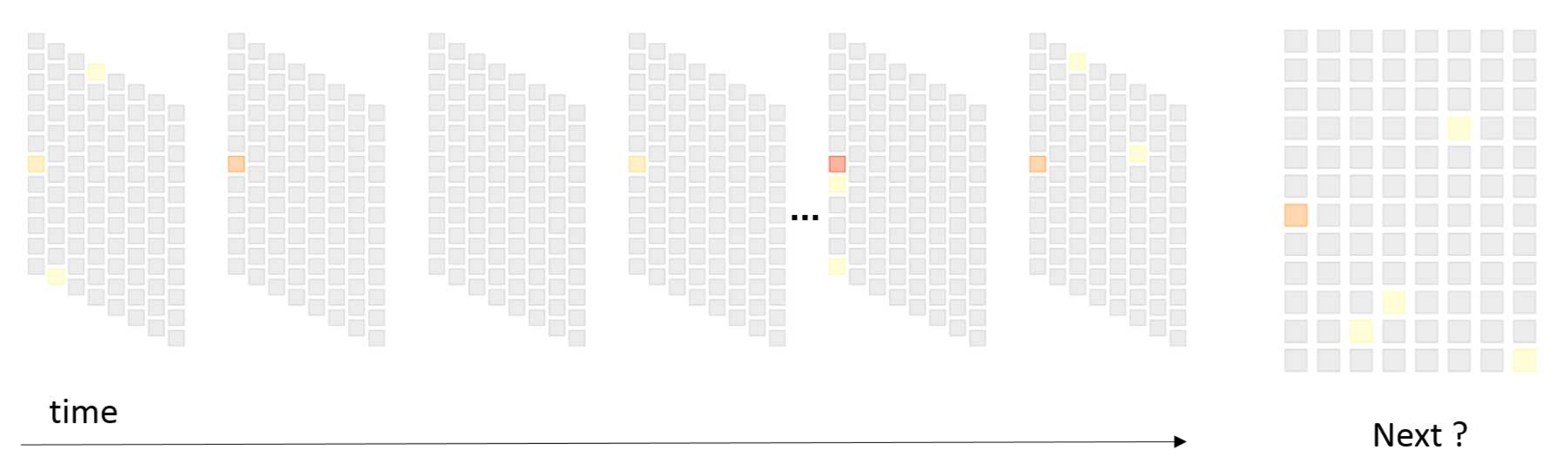}}
  \\
\vskip -0.1in
  \begin{tabular}{@{}c@{}}
  \subfloat[TDRN]{\includegraphics[width=0.45\linewidth,height=0.21\linewidth]{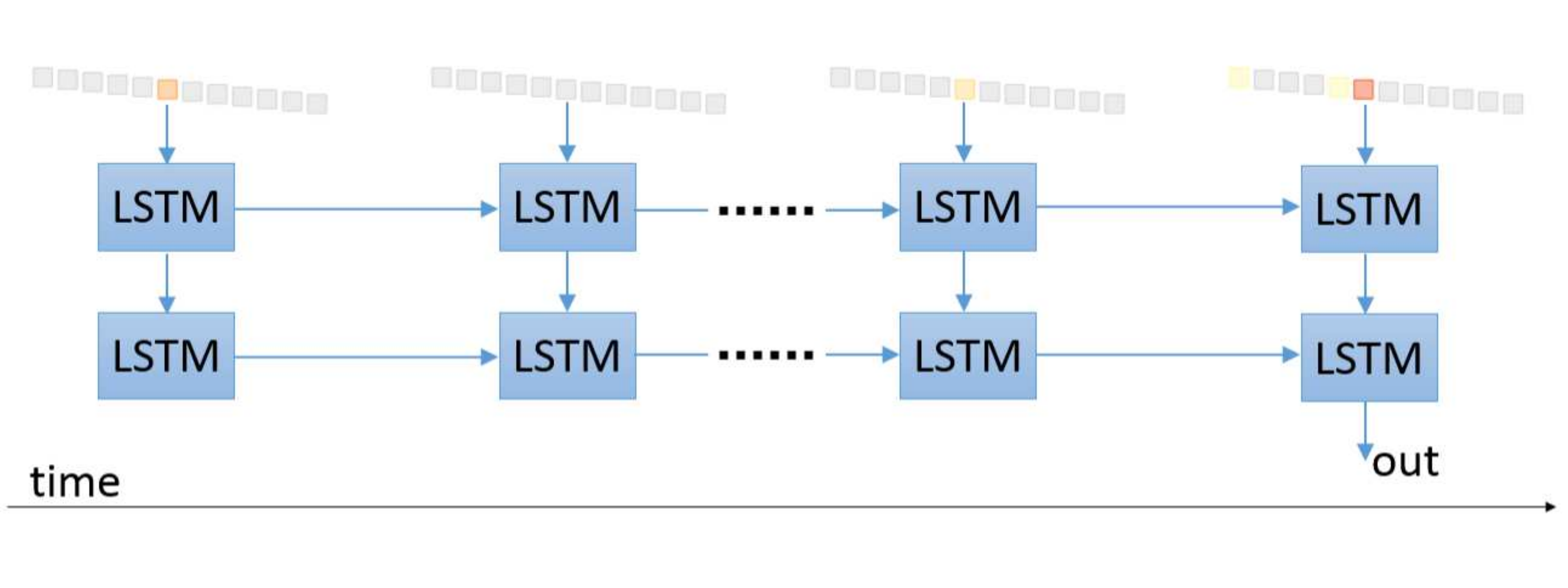}}
  \end{tabular}
  \begin{tabular}{@{}c@{}}
  \subfloat[LRCN]{\includegraphics[width=0.53\linewidth,height=0.34\linewidth]{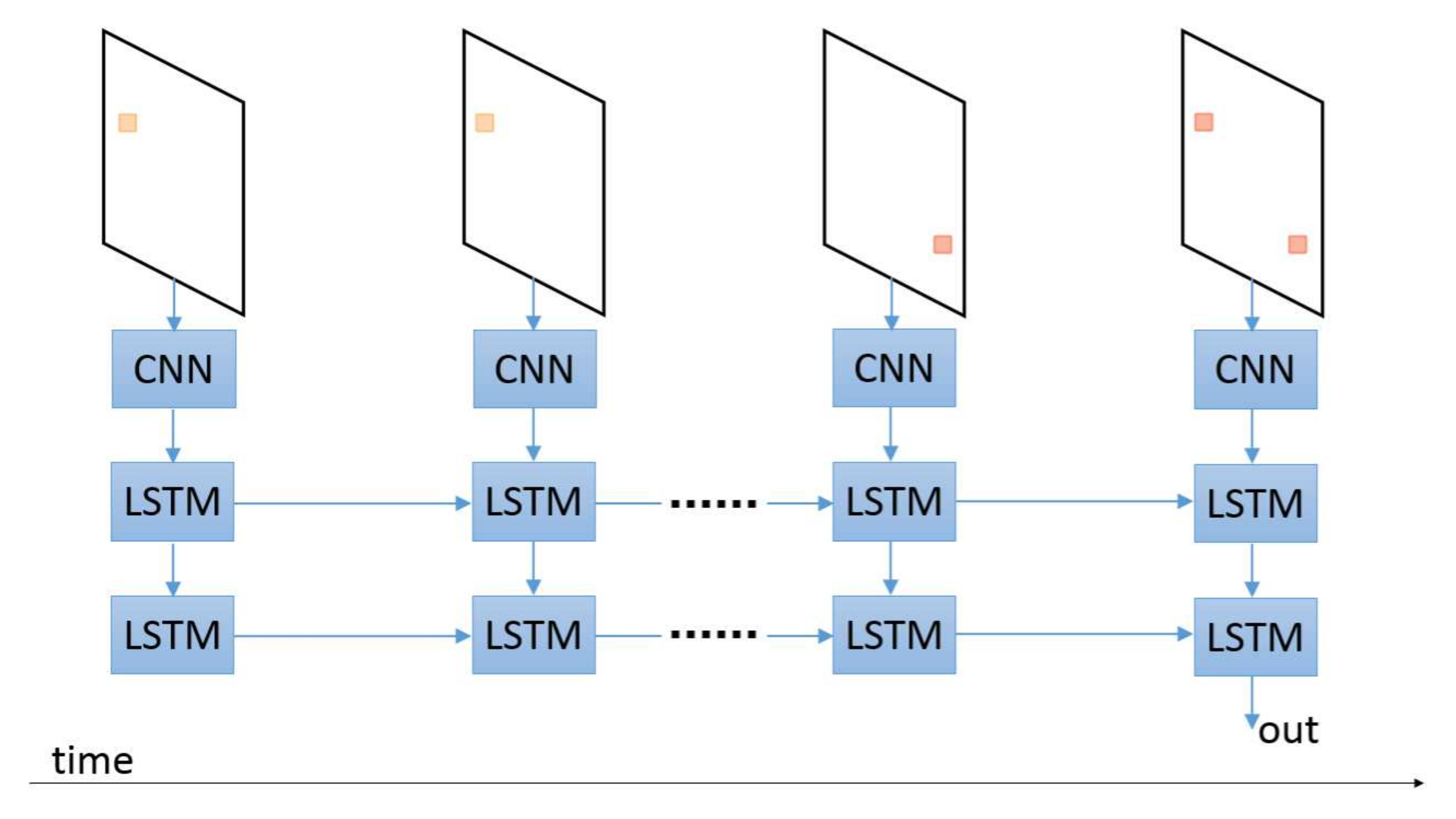}}
  \end{tabular}
\caption{Different learning architecture for topical behavior prediction.}
\label{fig-deep}
\vskip -0.15in
\end{figure}

In this work, we predict the response whose space is the same as the input space, using the historical behavioral data. The input data is the network access log containing the entity ID, the timestamp, and the meta data about the accessed resource. First, we organize the activities into topics. The topical activities on each topic is then quantified and measured for each entity. Over several periods of time, we observe the topical behavior over the same set of topics for all entities in the experiment. The prediction task is as Figure~\ref{fig-deep}(a). Several combinations of deep neural network are explored to predict topical behavior. Specifically, the RNN such as the long short-term memory units (LSTM) \cite{hochreiter97} is employed to learn the temporal variation patterns of the topical behavior. The CNN and the locally-connected network (LCN) \cite{lecun98} are used to learn the spatial composition of the topical behavior. The relationship between topics needs to be abstracted and evenly distributed like pixels for the CNN and the LCN to learn.

\begin{figure*}[tb]
\vskip -0.15in
  \centering
  \subfloat[]{\includegraphics[scale=0.155]{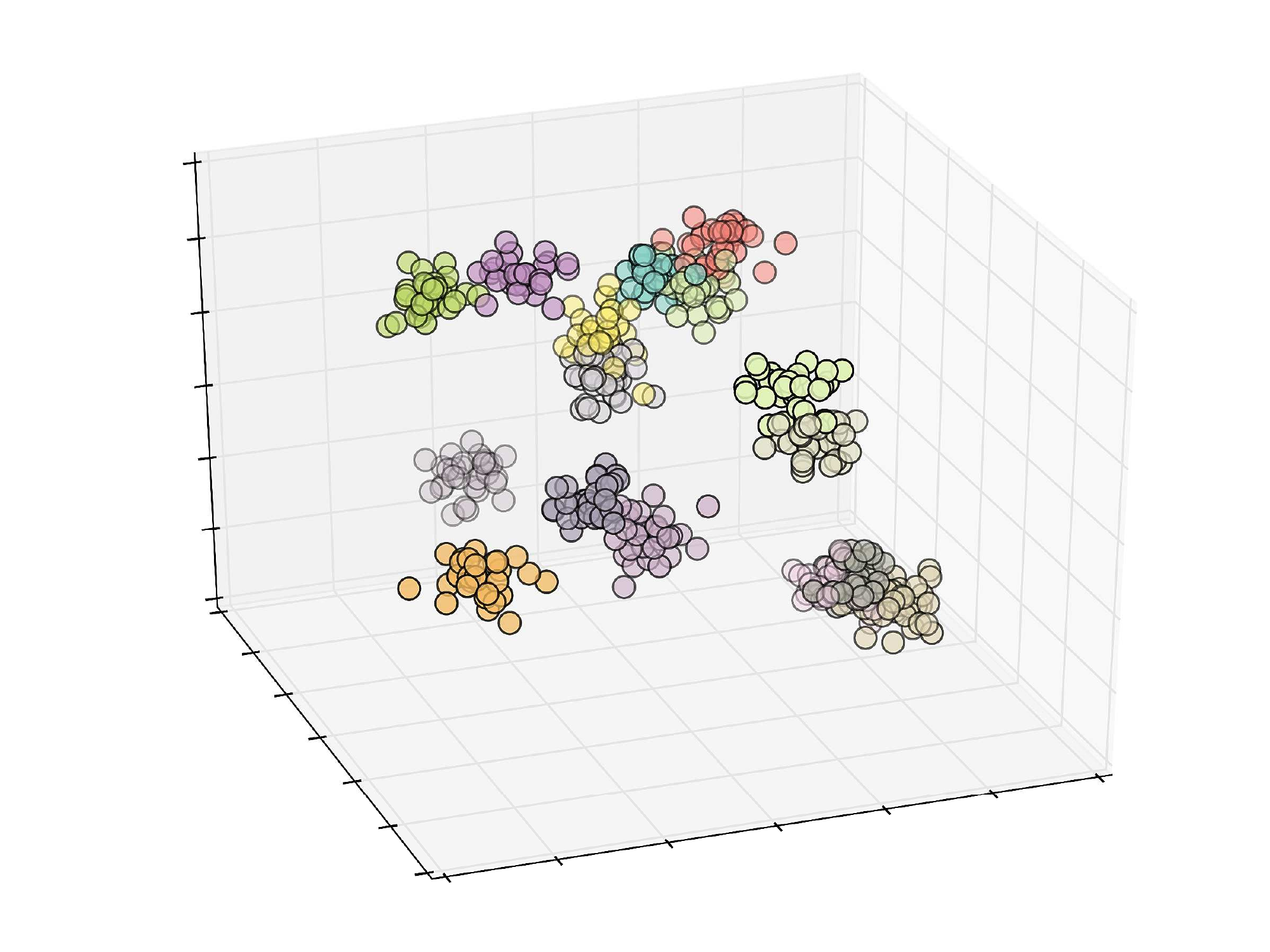}} \quad\quad
  \subfloat[]{\includegraphics[scale=0.145]{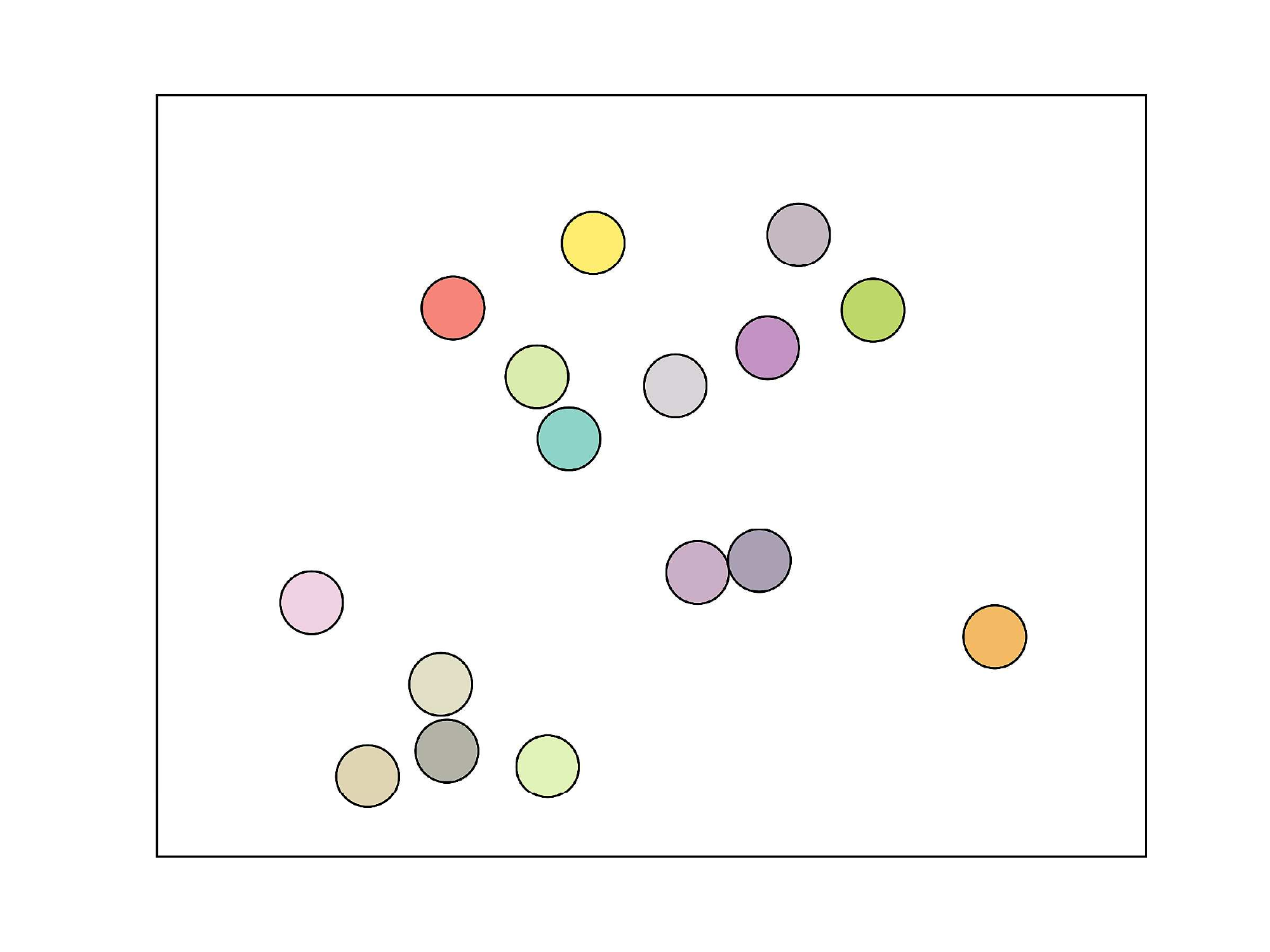}} \quad\quad
  \subfloat[]{\includegraphics[scale=0.145]{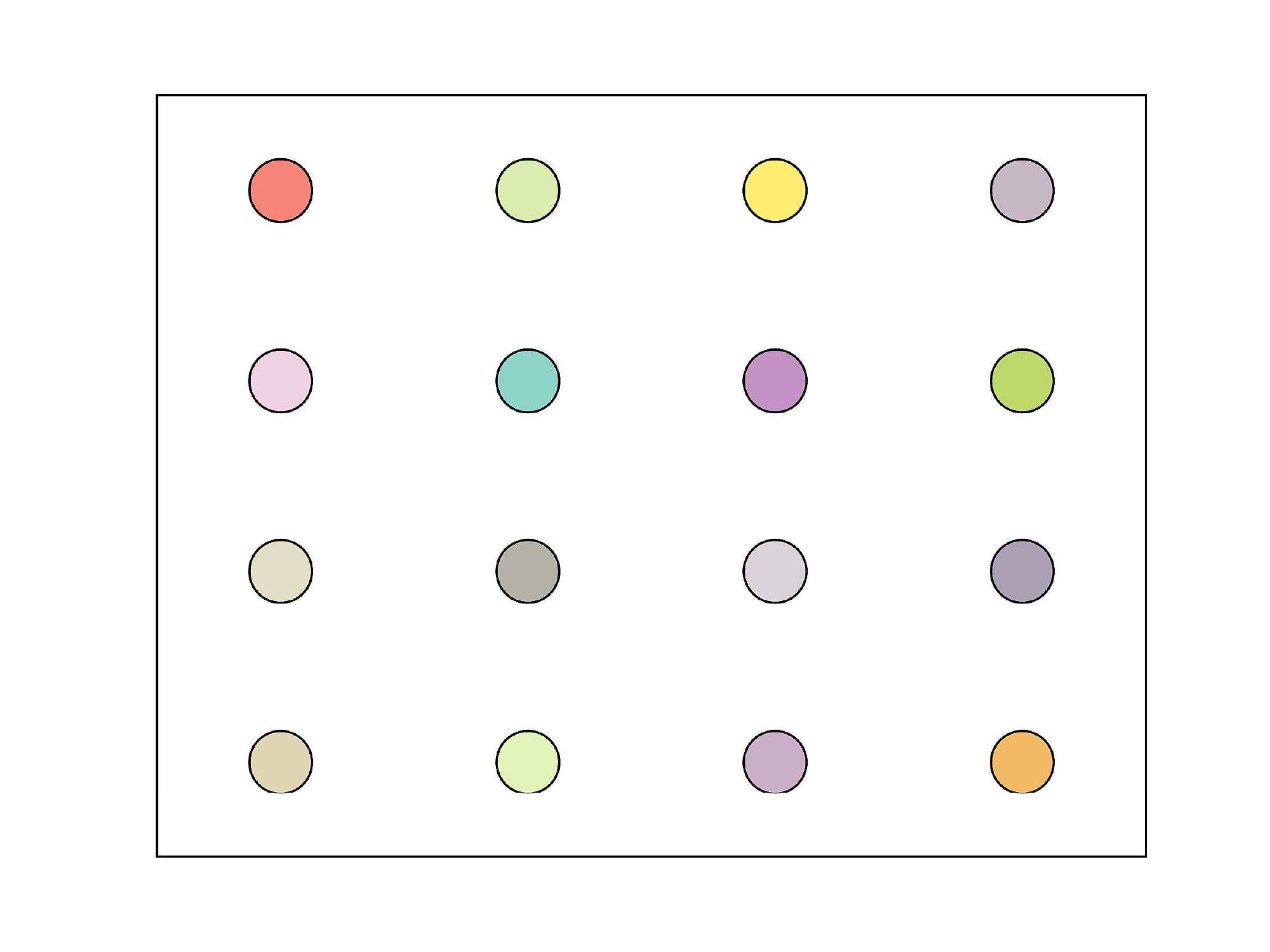}} \quad\quad
  \subfloat[]{\includegraphics[scale=0.023]{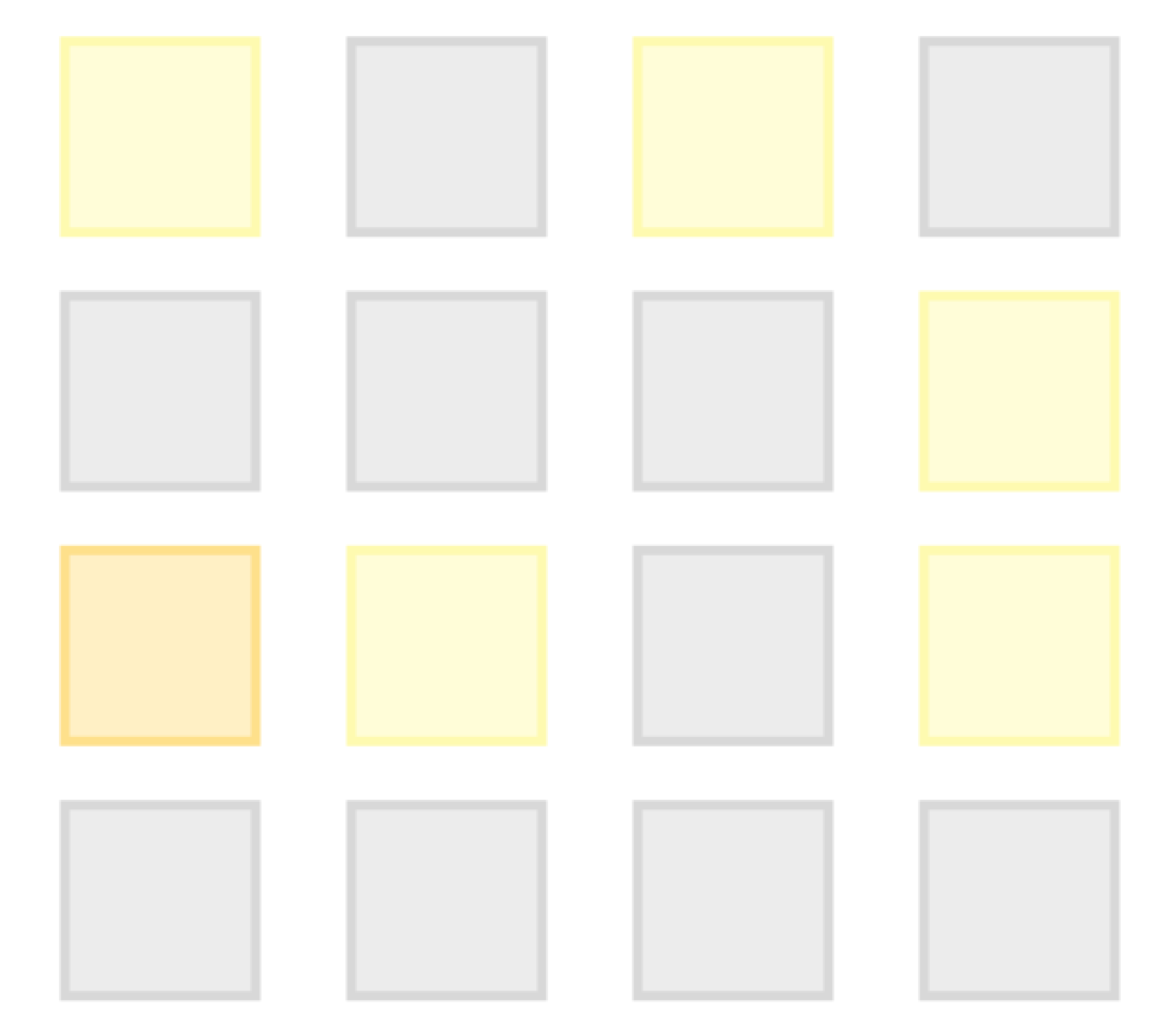}}
\caption{Topical behavior. (a) data points in the high dimensional space, where topics are extracted; (b) topics after dimension reduction; (c) topics after homogeneous mapping; (d) topical metrics for an entity}
\label{fig-TP}
\vskip -0.15in
\end{figure*}


\section{Method} \label{sec-method}

To keep the behavior prediction within a trackable scope, we summarize the input network activities into topics. Starting from the activity log of all entities over a benchmark period of time, the latent latent Dirichlet allocation (LDA) algorithm is applied to find the topics in a high dimension word vector space. For each entity, the vectorized log entries are summarized on these topics to form quantitative metrics. For example, the topical volume over topic $t$ of entity $e$ can be measured as
\begin{equation} \label{eq-vol}
V^{(B_{e,T})}_t = \log(\sum_{a\in B_{e,T}}r_a+1)),
\end{equation}
where $r_a$ is the relevancy for activity $a$ to topic $t$, and $B$ is the set of activities defined by the unique content documents of all activities logged within the time period $T$.

To better explored the intra-topic relationship in the behavioral data, we want to capture the co-occurrence detail between any pair of topics. Furthermore, we want to discriminate the detail according to the distance between the topic pair - the co-occurrence means more when the two topics are closer. In order to arrange the topical metrics to be similar to the pixels in an image into the CNN, the topical metrics need to go through the following two steps, as illustrated in Figure~\ref{fig-TP}.

\begin{enumerate}
  \item Dimension reduction: use methods like PCA or t-SNE to map the topical metrics into a 2D or 3D space, while maintaining the spatial relationship among topics.
  \item Homogeneous mapping: the topical metrics need to be placed evenly over the visualization space for the CNN to digest. The spatial relationship among the topics also needs to be maintained. One way to achieve this goal is the split-diffuse (SD) algorithm \cite{Su16}.
\end{enumerate}

\begin{figure*}[!h]
  \centering
  \subfloat[$RLE$ on training data]{\includegraphics[scale=0.11]{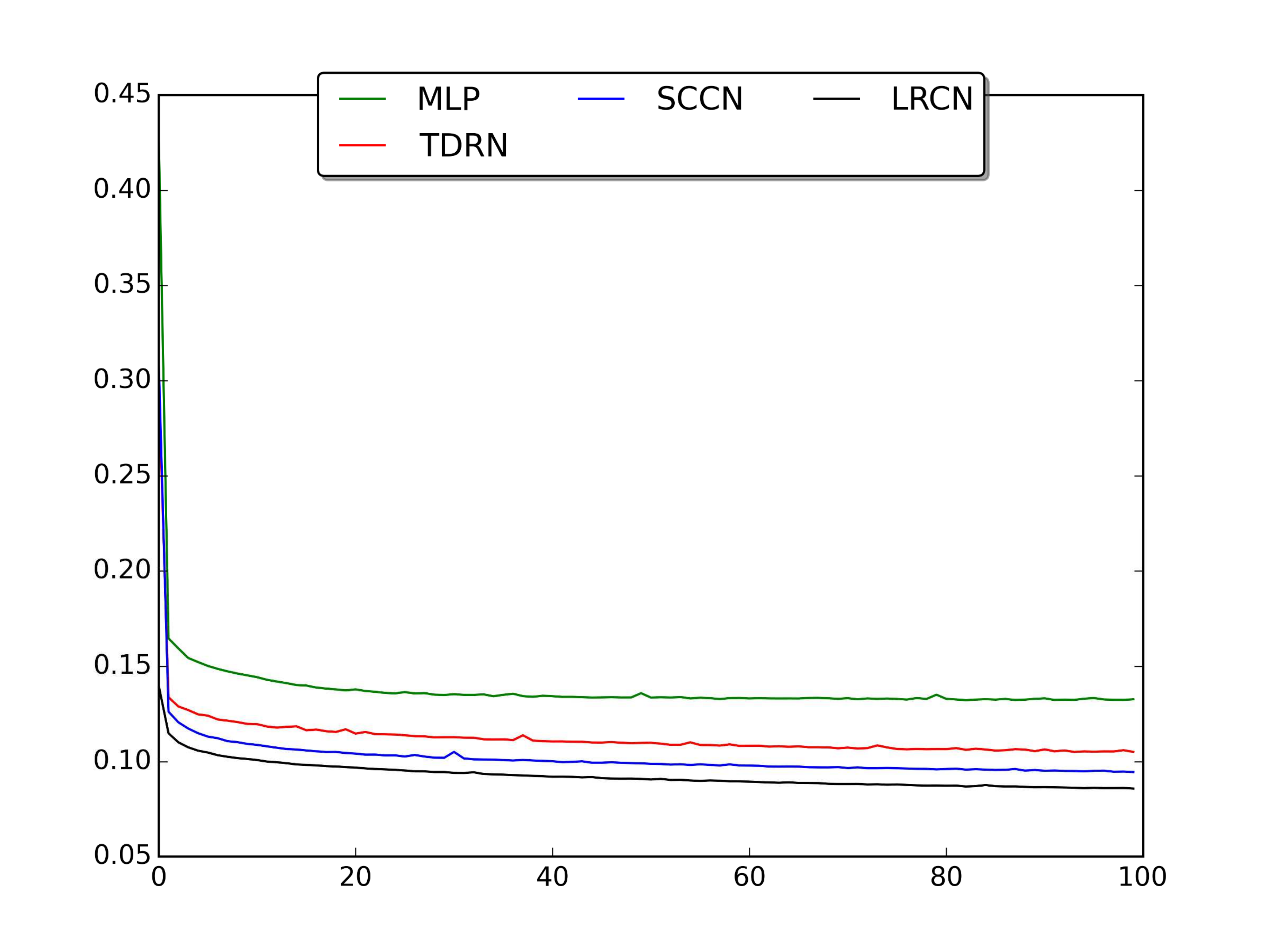}} \quad
  \subfloat[$RLE$ on validation data]{\includegraphics[scale=0.11]{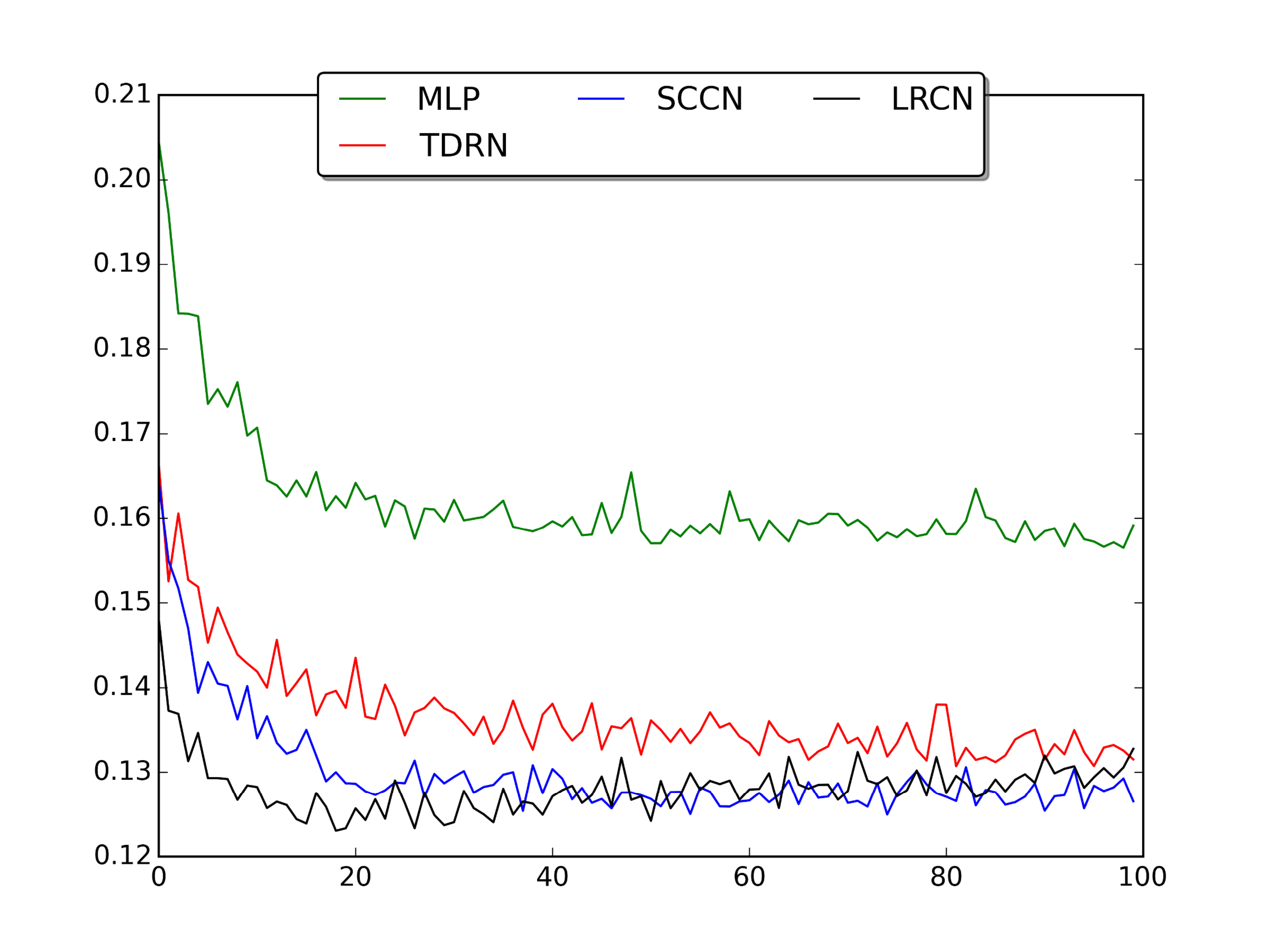}} \quad
  \subfloat[$RLE$ on testing data]{\includegraphics[scale=0.11]{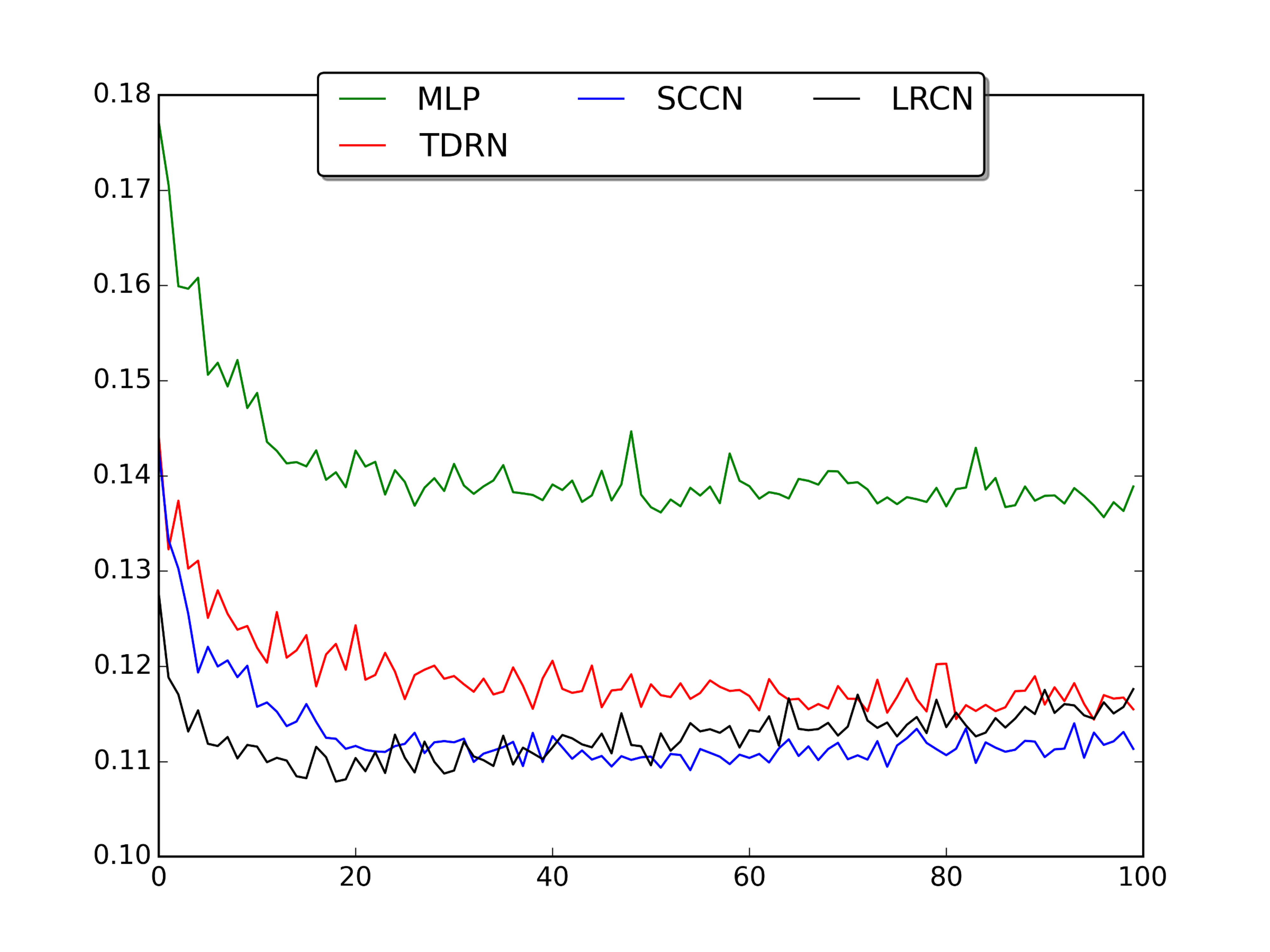}} \quad
  \subfloat{\mytab}
\caption{Performance on various learning architectures; x-axis indicates the number of epochs.}
\label{fig-R}
\vskip -0.1in
\end{figure*}

\section{Temporal and Spatial Learning}

We adopt various architectures to study how the temporal and the spatial information can help learning the topical behavior. For the benchmark MLP, the topical metrics over different time periods are cascaded into one single 1D vector for each sample. For the time distributed recurrent network (TDRN) in Figure~\ref{fig-deep}(b), we use one layer of LSTMs to track the topical metrics as a sequence for each time period, and then another layer of LSTMs to track the output states.

The long-term recurrent convolutional networks (LRCN~\cite{donahue15}) combines the convolutional layers with the long-range temporal recursion. The LRCN in Figure~\ref{fig-deep}(c) exploits both the temporal and the spatial relationship among topics. LCN is another option to explore spatially while being time efficient. In the proposed spatially connected convolutional networks (SCCN), the convolutional units in LRCN are replaced by the LCNs. The LCNs do not share the trained weights between different position. Instead, the same set of weights is applied to the same position across samples.

In predicting the trending or risky topic, the cost of missed future trend is higher than the cost of false positives. One candidate loss metric, the risk loss error (RLE), is defined as
\begin{equation} \label{eq-rle}
 RLE = \frac{1}{|\mathcal{V}|} \sum_{\forall v \in \mathcal{V}} {v(\hat{v}-v)^2},
\end{equation}

The data set comes from 150-million activity log entries of $98,881$ network entities, split into $69,407$ training, $7,712$ validation and $21,762$ testing samples. Each log entry contains the entity ID, timestamp, and the description of the accessed network resource. For any sample over any time period, the content of its log entries is summarized as in Equation~\ref{eq-vol} with the pretrained LDA from the benchmark log content. Each sample has the topical metrics over time as in Figure~\ref{fig-deep}. The predicted target values in the testing samples are from the time periods that are later then all the training time periods. This setup simulates the behavioral prediction for the future. Regulations are adopted to generalize the models.

\section{Result and Conclusion}

Figure~\ref{fig-R} shows the experimental result on loss metrics $RLE$. Experiments were also conducted on other metrics (the results being omitted). With only the temporal relationship explored by the TDRN, the prediction gain against the MLP ranging from $11.32\%$ to $16.73\%$ depending on the loss metric and the evaluation scenario. The LRCN explores both temporal relationship and spatial relationship over topics. The additional spatial information among topics tracked by CNN further improve the prediction gain from $13.73\%$ to $19.92\%$. Replacing the CNN spatial tracking with the LCNs, the SCCN provides a comparable $14.20\%$ to $19.85\%$ prediction gain to the LRCN. The locally customized patch dictionaries allows the SCCN to be better regulated than the LRCN.  Nevertheless, the SCCN is $1.5$ to $3$ times faster than the LRCN in the comparable setup. This is similar to the observation in \cite{chen12}.

In conclusion, a new summarization framework is proposed to predict the topical behavior for all network entities. Several state-of-the-art learning architectures are tested to verify the temporal- and the spatial-gain. Within this new framework, the proposed SCCN is more efficient and better regulated.

\bibliographystyle{IEEEtran}
\bibliography{IEEEabrv,jsu2016}

\end{document}